\documentclass{article} % For LaTeX2e
\usepackage[final]{colm2026_conference}

\usepackage{microtype}
\usepackage{hyperref}
\usepackage{amsmath}
\usepackage{cleveref}
\usepackage{url}
\usepackage{booktabs}

\usepackage{lineno}

\definecolor{darkblue}{rgb}{0, 0, 0.5}
\hypersetup{colorlinks=true, citecolor=darkblue, linkcolor=darkblue, urlcolor=darkblue}

\usepackage{booktabs}
\usepackage{tabularx}
\usepackage[table]{xcolor}
\usepackage{colortbl}
\usepackage{arydshln}
\usepackage{tcolorbox}
\usepackage{pifont}
\usepackage{fvextra} 
\DefineVerbatimEnvironment{VerbatimWrap}{Verbatim}{
breaklines=true, 
breakindent=0pt, 
breaksymbol={},
fontsize=\small,
}

\definecolor{darkblue}{rgb}{0, 0, 0.5}
\definecolor{chocolate}{HTML}{D2691E}
\definecolor{maroon}{HTML}{A00000}
\definecolor{indigo}{HTML}{4B0082}
\title{Procedural Knowledge at Scale Improves Reasoning}

\author{
Di Wu\textsuperscript{1}\thanks{Work done while at Meta FAIR.}
\quad Devendra Singh Sachan\textsuperscript{2}
\quad Wen-tau Yih\textsuperscript{2}
\quad Mingda Chen\footnotemark[1] \\
\textsuperscript{1}University of California, Los Angeles
\qquad \textsuperscript{2}Meta FAIR \\
\texttt{diwu@cs.ucla.edu}
}

\begin{document}

\ifcolmsubmission
\linenumbers
\fi

\maketitle

\begin{abstract}
Test-time scaling has emerged as an effective way to improve language models on challenging reasoning tasks. However, most existing methods treat each problem in isolation and do not systematically reuse knowledge from prior reasoning trajectories. In particular, they underutilize procedural knowledge: how to reframe a problem, choose an approach, and verify or backtrack when needed. We introduce \textbf{Reasoning Memory}, a retrieval-augmented generation (RAG) framework for reasoning models that explicitly retrieves and reuses procedural knowledge at scale. Starting from existing corpora of step-by-step reasoning trajectories, we decompose each trajectory into self-contained subquestion-subroutine pairs, yielding a datastore of 32 million compact procedural knowledge entries. At inference time, a lightweight in-thought prompt lets the model verbalize the core subquestion, retrieve relevant subroutines within its reasoning trace, and reason under diverse retrieved subroutines as implicit procedural priors. Across six math, science, and coding benchmarks, Reasoning Memory consistently outperforms RAG with document, trajectory, and template knowledge, as well as a compute-matched test-time scaling baseline. With a higher inference budget, averaged across models and tasks, it improves over no retrieval by 19.5\% and over the compute-matched baseline by 8.9\%. Ablation studies show that these gains come from two key factors: the broad procedural coverage of the source trajectories and our decomposition and retrieval design, which together enable effective extraction and reuse of procedural knowledge. Our experiment code is available at \url{https://github.com/facebookresearch/reasoning-memory}.
\end{abstract}

\section{Introduction}
\label{section:intro}

Test-time scaling, where additional inference-time compute is allocated per instance to improve accuracy, has quickly become a central paradigm for frontier reasoning tasks in math, science, and coding \citep{DBLP:conf/nips/Wei0SBIXCLZ22, DBLP:conf/iclr/0002WSLCNCZ23, DBLP:conf/nips/YaoYZS00N23, DBLP:journals/corr/abs-2408-03314, DBLP:conf/iclr/ChenLSZ24, DBLP:journals/corr/abs-2501-09891, DBLP:journals/corr/abs-2501-19393}. Recently, reasoning models such as OpenAI’s o1 series \citep{o1}, DeepSeek R1 \citep{dsr1}, and QwQ \citep{qwq} are explicitly trained to produce ``thinking'' tokens. At inference time, their performance scales up as they are allowed to ``think longer'' with a larger token budget.

Despite this progress, one limitation of current test-time scaling strategies is that they do not reuse knowledge across reasoning runs. Such knowledge includes both \textbf{factual knowledge}, such as definitions, formulas, and theorems, and more importantly \textbf{procedural knowledge}: how to reframe a problem, decompose it into subquestions, exploit structure, choose an approach, and verify or backtrack when needed. Discarding procedural knowledge from prior reasoning trajectories can hurt both accuracy and efficiency, as models are forced to repeatedly rediscover useful strategies and revisit unsuccessful lines of thought.

Retrieval-augmented generation (RAG) provides a natural framework for accumulating useful knowledge offline and incorporating it at test time. Recent work has begun to interleave retrieval with reasoning or to introduce general-purpose datastores that improve reasoning performance \citep{DBLP:journals/corr/abs-2501-05366, DBLP:journals/corr/abs-2503-05592, DBLP:journals/corr/abs-2507-01297}. However, these systems primarily improve access to general background knowledge, while leaving reasoning-specific procedural knowledge implicit. For difficult reasoning tasks, the relevant bottleneck is often not missing facts alone, but missing guidance on what subquestion to solve next and how to solve it. As a result, retrieved context is often only loosely aligned with the model’s current reasoning state. This weak alignment can be especially problematic for reasoning models, whose long reasoning traces may amplify irrelevant contexts. As we later show in our pilot study, standard document-level RAG can yield limited or even negative gains for reasoning models, suggesting the challenge of retrieving knowledge in a form well aligned with the problem and easy for reasoning models to leverage (\S\ref{section:results-pilot}). High-quality procedural knowledge is rare in standard web corpora, and generic retrieval often fails to surface the right subroutine when it is needed.

\begin{figure}
    \centering
    \includegraphics[width=\linewidth]{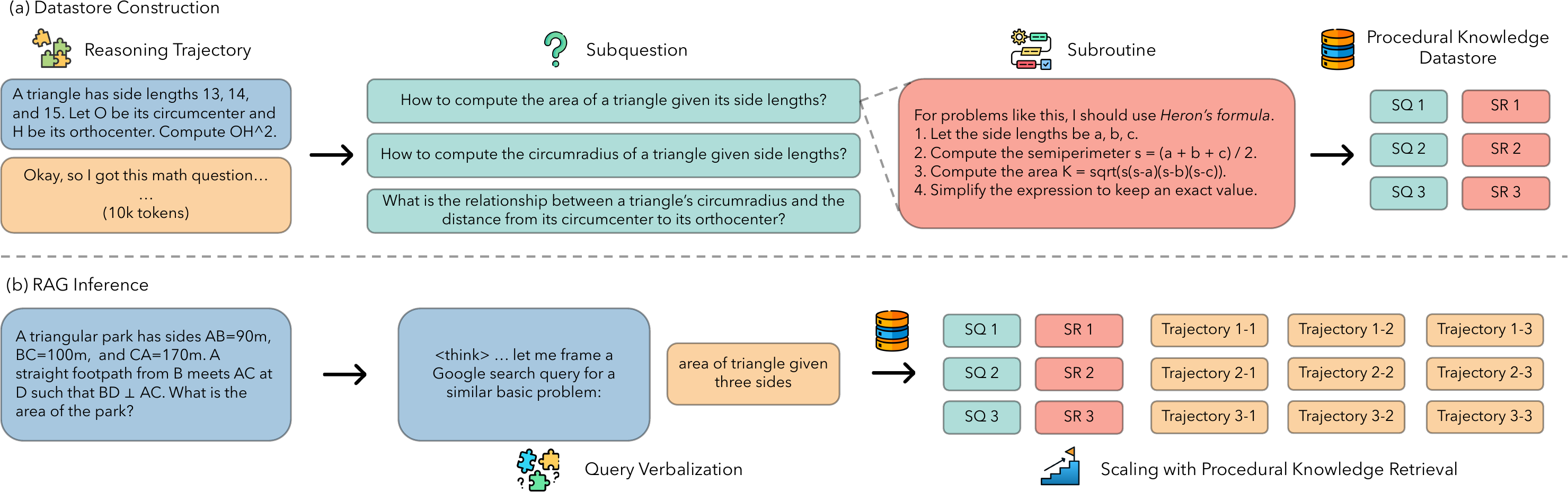}
    \caption{\textbf{Illustration of the Reasoning Memory framework.} (a) We extract self-contained procedural knowledge from diverse public reasoning trajectories to construct a datastore. (b) At inference time, we retrieve top-$k$ subquestion-subroutine pairs, generate separate groups of candidate trajectories under each subroutine, and filter and rank the candidates. The source trajectories are not inserted.}

    \label{fig:main-figure}
    \vspace{-4mm}
\end{figure}

To address this limitation, we introduce \textbf{Reasoning Memory}, a RAG framework that reuses procedural knowledge at scale for reasoning models. As demonstrated in \Cref{fig:main-figure}, rather than indexing generic documents, Reasoning Memory builds a procedural datastore from existing reasoning trajectories. We segment trajectories into atomic subquestion descriptions paired with concise high-level subroutines, yielding a datastore of over 32M items. At inference time, a lightweight in-thought query prompt encourages the model to verbalize its current subquestion as a short query. We retrieve relevant subquestion-subroutine pairs and inject them into the reasoning trace as implicit procedural priors; their source trajectories are not included. We then sample separate groups of reasoning trajectories under diverse retrieved procedures and apply a simple length-based uncertainty heuristic to filter both subroutines and candidate solutions before selecting the final answer. The result is a simple test-time scaling procedure in which additional inference budget enables broader exploration under diverse retrieved procedures, leading to stronger performance as compute increases.

Empirically, Reasoning Memory delivers consistent gains across open-weight reasoning models, inference budgets, and frontier benchmarks in math, science, and coding. It outperforms document-level RAG, trajectory- and template-based procedural RAG, and compute-matched test-time scaling without retrieval. With the higher inference budget, averaged across models and tasks, it improves accuracy by 19.5\% over no retrieval and by 8.9\% over the compute-matched baseline. Our analyses further show that performance improves with larger and more diverse procedural datastores, and under a fixed inference budget, it is more effective to explore multiple trajectories under a small set of high-quality retrieved subroutines than to spend the same compute under a single subroutine. Additional ablations on datastore construction and in-thought querying identify the contribution of our design.

\section{Related Work}
\label{section:related-work}
\paragraph{Test-time Scaling for Reasoning}
Test-time scaling allocates additional inference-time compute per instance to improve performance on challenging math, science, and coding tasks \citep{DBLP:conf/nips/Wei0SBIXCLZ22, DBLP:conf/iclr/ChenLSZ24, DBLP:journals/corr/abs-2412-19437}. Starting from Chain-of-Thought prompting \citep{DBLP:conf/nips/Wei0SBIXCLZ22}, prior work has explored two broad paradigms. \textit{Sequential scaling} extends a single trajectory through planning, reflection, self-correction, and backtracking \citep{DBLP:conf/nips/MadaanTGHGW0DPY23, DBLP:conf/iclr/ChenLSZ24, DBLP:journals/corr/abs-2501-09891, DBLP:journals/corr/abs-2501-19393, DBLP:journals/corr/abs-2505-05315}, while \textit{parallel scaling} samples multiple candidates and combines them through selection or aggregation \citep{DBLP:conf/iclr/0002WSLCNCZ23, DBLP:conf/nips/YaoYZS00N23, DBLP:journals/corr/abs-2408-03314}. Recent reasoning models explicitly trained to produce long thinking prefixes, including OpenAI's o1 series \citep{o1}, DeepSeek R1, and QwQ \citep{dsr1, qwq}, exhibit strong inference-time scaling behavior, and simple budget forcing can already yield strong gains on math and science benchmarks \citep{DBLP:journals/corr/abs-2501-19393}. Another line of work combines parallel scaling with search, including process reward models and Monte Carlo Tree Search style methods \citep{DBLP:conf/acl/WangLSXDLCWS24,DBLP:journals/corr/abs-2410-01707,DBLP:conf/naacl/ParkLGC25}. Our work targets reasoning models and augments test-time scaling with a  procedural knowledge datastore derived from prior reasoning trajectories. It supports sequential scaling by injecting retrieved procedures into the thinking process and parallel scaling by exploring different retrieved subroutines.
\paragraph{Memorizing Procedural Knowledge}
A complementary line of work studies how to represent and reuse procedural knowledge such as workflows, strategies, or templates distilled from experience. Agent Workflow Memory \citep{DBLP:journals/corr/abs-2409-07429} induces reusable workflows from web navigation traces, while ReasoningBank \citep{DBLP:journals/corr/abs-2509-25140} converts successful and failed episodes into compact strategies and supports memory-aware test-time scaling. Think-in-Memory \citep{think-in-memory} and Dynamic Cheatsheet \citep{dynamic-cheatsheet} maintain or adapt memories across interactions and tasks, while Retrieval-of-Thought \citep{retrieval-of-thought} and trajectory-informed memory \citep{trajectory-informed-memory} reuse prior thoughts or agent experience. Outside web agents, Buffer of Thoughts \citep{DBLP:conf/nips/YangYZCXZG024} distills high-level thought templates from problem-solving traces and retrieves them for new instances. Related approaches, including Self-Discover \citep{DBLP:conf/nips/ZhouP0CCLCZMZ24}, ReasonFlux \citep{DBLP:journals/corr/abs-2502-06772}, and RLAD \citep{DBLP:journals/corr/abs-2510-02263}, similarly operate over libraries of heuristics or natural-language abstractions to improve reasoning. Stylistic structure can itself affect reasoning distillation \citep{style-over-substance}, so we treat the plan-like form of subroutines as a possible aid rather than attributing gains to style alone. Our contribution is to operationalize procedural memory as a 32M-entry datastore, active in-thought retrieval interface, and compute-matched scaling recipe for reasoning models on math, science, and coding tasks.

\paragraph{RAG for Reasoning}
Retrieval-augmented generation (RAG) equips language models with external knowledge through retrieval and in-context conditioning \citep{DBLP:journals/corr/abs-2002-08909, DBLP:conf/nips/LewisPPPKGKLYR020, DBLP:conf/naacl/ShiMYS0LZY24}. While most RAG systems are developed for factuality-related tasks, a smaller line of work studies RAG for reasoning-intensive settings. CompactDS builds a compact and diverse web-scale datastore and applies standard retrieval to improve performance on multiple reasoning benchmarks \citep{DBLP:journals/corr/abs-2507-01297}. ReasonIR trains a retriever specialized for reasoning-intensive retrieval \citep{DBLP:journals/corr/abs-2504-20595}. TRACE constructs reasoning chains from retrieved evidence \citep{trace-evidence}, while RAT iteratively revises reasoning steps using retrieved information \citep{rat}. Agentic search systems such as Search-o1 \citep{DBLP:journals/corr/abs-2501-05366} and R1-Searcher \citep{DBLP:journals/corr/abs-2503-05592} instead let models issue queries during reasoning and incorporate retrieved documents. These methods primarily use retrieval to ground reasoning in external information. Reasoning Memory instead retrieves compact procedures distilled from prior reasoning trajectories and supplies them directly to the thinking process.

\section{Reasoning Memory: Approach}
\label{section:approach}

We introduce \textbf{Reasoning Memory}, a RAG framework that induces, retrieves, and reuses \emph{procedural} knowledge at scale for reasoning models. As shown in \Cref{fig:main-figure}, we first distill public reasoning trajectories into a datastore of \emph{subquestion-subroutine} pairs (\S\ref{section-approach-datastore}). At inference time, the model verbalizes its current subquestion as a compact query, and the retrieved procedures are injected directly into its thinking stream (\S\ref{section-approach-retrieval}). We then use multiple retrieved subroutines as implicit priors for test-time scaling (\S\ref{section-approach-scaling}).

\subsection{Procedural Knowledge Datastore}
\label{section-approach-datastore}

Problem-solving trajectories of current reasoning models contain rich procedural knowledge, but they are too long and noisy to reuse directly. Reasoning Memory therefore converts them offline into a \emph{procedural knowledge datastore} of natural-language subquestions and concise solution subroutines, which supports both effective retrieval and efficient in-context use.

Formally, let a trajectory instance be $(\mathbf{q}, \mathbf{a}, \mathbf{r})$, where $\mathbf{q}$ is the original problem, $\mathbf{a}$ is the final answer, and $\mathbf{r}$ is the intermediate reasoning trace. We map $(\mathbf{q}, \mathbf{r})$ to a set of $K$ subquestion-subroutine pairs $\{(q_i, s_i)\}_{i=1}^K$ using a two-step prompting pipeline. First, we derive self-contained subquestions $\{q_i\}$ that capture the key intermediate goals in the trajectory. Then, for each $q_i$, we generate a concise subroutine $s_i$ that summarizes the high-level procedure used to address it, while abstracting away local calculations and incidental trial-and-error. We do not enforce a single abstraction level: the datastore contains both broad strategies and local procedures to match subquestions of different granularity. This yields a reusable representation that remains short enough to serve as an in-context procedural prior. A concrete example is shown in \Cref{fig:example-datastore-content} in the appendix.

Rather than collecting trajectories from scratch, we build the datastore from the publicly released Nemotron V1 corpus \citep{NemotronPostTrainingDatasetV1}, which covers diverse math, science, and coding questions. The resulting datastore contains approximately 32 million $(q_i, s_i)$ pairs. On average, subquestions contain 19.2 tokens, subroutines 207.9 tokens, and each trajectory contributes 10.5 subquestions. The full datastore construction details are presented in \Cref{appendix-reasoning-memory-impl-details}. QwQ-32B is used for both prompting steps\footnote{Smaller models are also effective, as most of the knowledge is already in the trajectories (\Cref{tab:ablations-other}).}.

\subsection{In-Thought Active Retrieval Augmentation}
\label{section-approach-retrieval}

Given a large datastore of self-contained problem-procedure pairs, the next challenge is querying it effectively to benefit a reasoning model's thinking process. To begin with, prompting the model to follow certain query generation formats within its thoughts is challenging. On the other hand,  approaches such as Search-o1 \citep{DBLP:journals/corr/abs-2501-05366} that defer query generation until extensive reasoning is expensive and is better suited factuality-related gaps than reasoning strategies which require early interventions to model thinking.

Inspired by \citet{active-rag}, we instead leverage a simple \emph{thought-hijacking} prompt to enable query generation part of the thinking stream itself. Concretely, we start the model's thinking with the following meta-thinking sentence:

\begin{center}
    \texttt{Now, let me search for a similar basic problem whose solution \\ can help unblock me for solving the current step. Let me\\ frame it as a more high-level google search query: }
\end{center}

We extract the next sentence in the continuation as the retrieval query $\tilde{q}$. Here, ``search query'' denotes a short search-style verbalization rather than a keyword-only query. It identifies the initial subproblem and removes incidental details such as long specifications, ad-hoc entities, symbols, examples, and formatting constraints. In practice, this produces concise subquestion descriptions that align well with datastore subquestions. Given $\tilde{q}$, we retrieve the top-$k$ subquestions and their corresponding subroutines, denoted by $\{(\hat{q}_j, \hat{s}_j)\}_{j=1}^k$, using a standard dense retriever. The subquestion acts as a concise key for its associated procedure; matching a question-like query to this key is more direct than matching it to procedural prose. The original source trajectories are not inserted at inference time. For each retrieved pair, we insert the following hint directly into the reasoning stream:

\begin{center}
    \texttt{[hint] Here is a problem solving procedure for a related \\question }``$\hat{q}_j$"\texttt{: }$\hat{s}_j$\texttt{ [end of hint]}
\end{center}

This hint is followed by a simple continuation cue, ``\texttt{Okay,}''. The model then continues reasoning conditioned on the retrieved procedural prior. Because both the query and the retrieved subroutine appear in the same thinking channel, retrieval acts as a lightweight extension of the model's ongoing reasoning.

\subsection{Inference-Time Scaling with Reasoning Memory}
\label{section-approach-scaling}

Finally, Reasoning Memory introduces a test-time scaling method that naturally adapts to different compute budgets (in terms of samples) by adjusting the diversity of retrieved procedures and intensity of per-procedure compute. Given a sampling budget of at most $m$ samples and top-$k$ retrieved subroutines $\{(\hat{q}_j, \hat{s}_j)\}_{j=1}^k$, we allocate $\lfloor m/k \rfloor$ samples to each retrieved subroutine using the retrieval-augmented prompt from \S\ref{section-approach-retrieval}. Each subroutine conditions a separate group of newly generated trajectories rather than being concatenated with the other retrieved subroutines. This yields a pool of candidate trajectories $\{\pi_{j,\ell}\}$, where $j \in \{1,\dots,k\}$ indexes the retrieved subroutine and $\ell \in \{1,\dots,\lfloor m/k \rfloor\}$ indexes samples under that subroutine.

We then score each trajectory $\pi_{j,\ell}$ with an uncertainty measure $r_{j,\ell}$ and normalize scores across the full pool:
\[
\tilde r_{j,\ell} \equiv \frac{\max_{j',\ell'} r_{j',\ell'} - r_{j,\ell}}{\max_{j',\ell'} r_{j',\ell'}-\min_{j',\ell'} r_{j',\ell'} }\,.
\]
For each retrieved subroutine, we compute an average quality score
\[
\bar{r}_j \equiv \frac{1}{\lfloor m/k \rfloor}\sum_{\ell=1}^{\lfloor m/k \rfloor}\tilde r_{j,\ell}\,.
\]
We retain subroutines with $\bar{r}_j > \tau$, rank their associated trajectories by confidence, and keep the top $n$ samples for pass@1-based evaluation. If fewer than $n$ samples remain, we backfill with the highest-confidence samples from the filtered-out pool.

Inspired by \citet{DBLP:journals/corr/abs-2505-17813}, we use the thinking length as the default uncertainty score. The intuition is that uncertain reasoning tends to produce longer traces due to extra branching and backtracking. Concretely, we set $r_{j,\ell}$ to the generated trajectory length in tokens, so that after normalization, larger $\tilde r_{j,\ell}$ indicates a shorter and potentially more confident trajectory relative to the rest of the pool. The subroutine score $\bar{r}_j$ therefore measures whether a retrieved subroutine tends to induce shorter samples. By design, the pipeline is compatible with other uncertainty signals, such as log-likelihood, entropy, or self-evaluated relevance. Nevertheless, we find length to be a strong and simple choice among these alternatives and use it for the main results. Length is only a heuristic confidence proxy: verbosity differences and premature convergence can break the correlation between shorter traces and correctness. Averaging across samples for each subroutine reduces reliance on any single short trajectory, but does not eliminate mismatched retrievals (\S\ref{appendix-qualitative-study}). We provide a full comparison in \S\ref{appendix-uncertainty-metrics}.

\section{Experimental Setup}
\label{section:exp-setup}

\paragraph{Models}
We focus on open-weight reasoning models that expose their thinking tokens. Our main evaluation uses three models spanning different sizes, base model families, and post-training recipes: DeepSeek-R1-Distill-Llama-8B \citep{dsr1}, OpenThinker3-7B \citep{ot3}, and Qwen3-32B \citep{qwen3}. Each model is used in the reasoning mode without additional fine-tuning.

\paragraph{Benchmarks and Metrics}
We evaluate on challenging math, science, and coding benchmarks. For math, we use AIME 2024 and AIME 2025 \citep[30 problems each;][]{aime2024}, and MATH500 \citep{math500}, a standard subset of the MATH test set \citep{math}. For science, we use GPQA-Diamond (GPQA-D) \citep{gpqa}. For coding, we use LiveCodeBench (LCB) \citep{livecodebench}; following the filtering criteria of \citet{DBLP:journals/corr/abs-2501-05366}, we evaluate on 112 problems from releases V1--V4 and 109 problems from releases V5--V6. We use standard metrics from prior work: math-equal for AIME and MATH500, exact match for GPQA-D, and execution-based pass@1 for LiveCodeBench, using the implementations of \citet{DBLP:journals/corr/abs-2501-05366}. Unless otherwise specified, we sample $m=8$ trajectories per problem and report the averaged performance (equivalent to pass@1). For methods with uncertainty-based selection, the performance is averaged over the selected $n$ samples after filtering. Our main setting uses $(m,n)=(8,4)$. We also evaluate $(m,n)=(30,8)$ and analyze larger budgets up to $m=100$ in \S\ref{section:results-analysis-scaling}.

\paragraph{Baselines}
We compare Reasoning Memory against baselines covering RAG with trajectory-level knowledge, template-based procedural knowledge, document-level factual knowledge, as well as retrieval-free uncertainty-based test-time scaling.

\begin{itemize}
\item \textbf{No RAG.} The model solves each problem without retrieval.

\item \textbf{Trajectory RAG.} We retrieve reasoning trajectories from the Nemotron v1 corpus using the question as the query, and either (1) prepend the first 256 tokens of each of the top 3 trajectories (\textit{prefix}) or (2) compress the top trajectory into a short summary and prepend it (\textit{summary}). This baseline uses the same source corpus as Reasoning Memory, but operates at the trajectory level rather than the subroutine level.

\item \textbf{Template RAG.} We instantiate RAG over small human-designed or automatically distilled template libraries from prior work. The \textit{ReasonFlux} variant uses templates from \citet{reasonflux}, and the \textit{Self-Discover} variant uses templates from \citet{DBLP:conf/nips/ZhouP0CCLCZMZ24}. For fairness, all retrieved reasoning templates are injected into the thinking stream in the same way as in Reasoning Memory.

\item \textbf{Document RAG.} We include two factual knowledge retrieval baselines: retrieving from \textit{CompactDS} following \citet{DBLP:journals/corr/abs-2507-01297}, and retrieving from Google\footnote{We set the search date cutoff to 2023/12/31 to prevent data contamination.}. In both cases, the original question is used as the query and the retrieved passages are prepended to the question in the prompt, following \citet{DBLP:journals/corr/abs-2501-05366}.

\item \textbf{Length Scaling.} A retrieval-free test-time scaling baseline inspired by \citet{DBLP:journals/corr/abs-2505-17813}. We sample $m$ independent trajectories and select $n$ high-confidence ones using the length-based uncertainty heuristic from \S\ref{section-approach-scaling}. This isolates the benefit of procedural retrieval\footnote{In the main text, we do not apply length scaling to the other baselines, since this is not standard practice. When added, the relative performance trends remain similar. Results are reported in \S\ref{appendix-baselines-with-len-scaling}.}. We evaluate $(m,n)=(8,4)$ and $(30,8)$.

\item \textbf{Reasoning Memory.} Our full method retrieves $k$ procedural subroutines from the Nemotron-based datastore, injects them in thought using the thought-hijacking prompt, and performs two-stage length-based filtering over $m$ trajectories. We use $\tau=0.1$ and evaluate $(m,n,k)$ = $(8,4,3)$ or $(30,8,10)$, carefully matching the sampling budget of the Length Scaling baseline.
\end{itemize}

For all retrieval-based methods, we use ReasonIR-8B \citep{DBLP:journals/corr/abs-2504-20595} as the retriever. Full implementation and efficiency details are provided in \S\ref{appendix-impl-details}.

\section{Results}
\label{section:results}

\subsection{Standard Document RAG is Poorly Aligned with Reasoning Models}
\label{section:results-pilot}

We begin with a pilot study to test whether a standard document-level RAG pipeline benefits reasoning models in the same way it has been shown to benefit instruction-tuned models. CompactDS reported that a web-scale general document datastore can improve instruction-tuned models on challenging reasoning benchmarks, but did not evaluate this setup on reasoning models \citep{DBLP:journals/corr/abs-2507-01297}. We therefore follow the CompactDS setup and compare paired instruction-tuned and reasoning models from the same families on AIME 2024, GPQA-Diamond, and LiveCodeBench. Full setup details are presented in \Cref{appendix-pilot-study}.

\begin{figure}[t]
\vspace{-4mm}
\centering
\includegraphics[width=\linewidth]{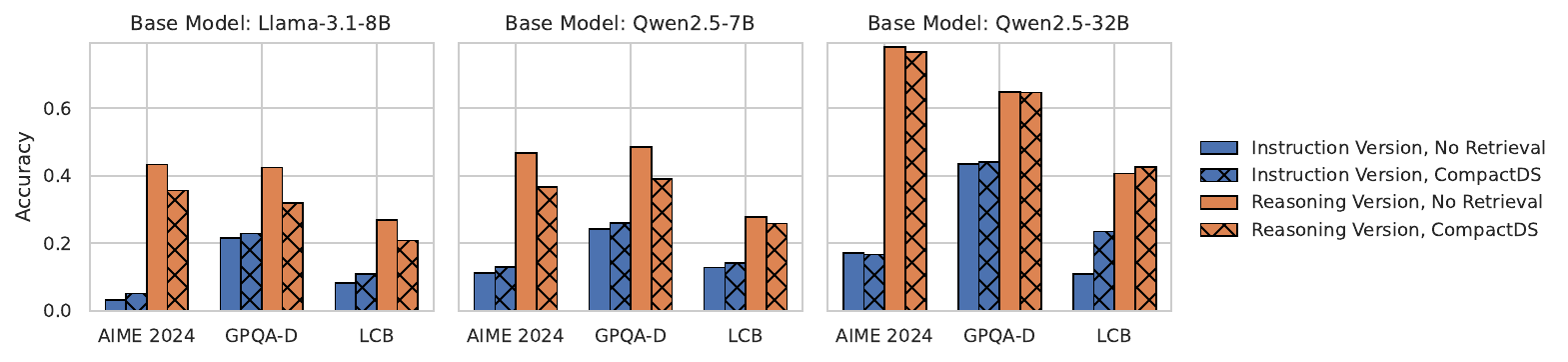}
\caption{\textbf{Standard document RAG benefits instruction-tuned models more than reasoning models.} Under the CompactDS pipeline, instruction-tuned models obtain modest gains from retrieval, whereas the corresponding reasoning models often see limited gains or even degradation, despite much stronger no-retrieval performance.}
\label{fig:compactds-instr-reason}
% \vspace{-4mm}
\end{figure}

\Cref{fig:compactds-instr-reason} shows a consistent pattern across model families and tasks: instruction-tuned models benefit modestly from retrieved documents, while the corresponding reasoning models often gain little or even degrade. This suggests that standard document-level RAG is poorly aligned with reasoning models, whose current subquestions often require procedural guidance rather than generic background context. In \Cref{appendix-pilot-knowledge}, we further diagnose this phenomenon using synthesized knowledge. The results show that reducing retrieval noise makes injected knowledge more effective, and that the \textit{type} of knowledge also matters: procedural guidance yields larger gains than factual knowledge for reasoning models. This distinction is functional rather than absolute: a subroutine may include a fact, theorem, or formula, but explains how to apply and verify it within a solution process. This motivates our focus on compact procedural subroutines rather than general documents.

\subsection{Main Results}
\label{section:results-main}

\newcommand{\dingup}[1]{\textsuperscript{\ding{#1}}}

\begin{table*}[t]
\centering
\resizebox{\linewidth}{!}{
\begin{tabular}{l l l ccc c cc}
\toprule
\textbf{Method} & \textbf{Ret.} & $m$ &
\multicolumn{3}{c}{\textbf{Math}} & \textbf{Science} &
\multicolumn{2}{c}{\textbf{Code (LCB)}} \\
 &  &  & \textbf{AIME 24'} & \textbf{AIME 25'} & \textbf{MATH500} & \textbf{GPQA-D} & \textbf{V1-4} & \textbf{V5-6} \\
\midrule
\rowcolor{gray!20} 
 \multicolumn{9}{c}{\textbf{DeepSeek-R1-Distill-Llama-8B}} \\
No RAG & N/A & 8 & 0.461 & 0.350 & 0.806 & 0.427 & 0.220 & 0.211 \\
Trajectory RAG (Prefix) & Proc. & 8 & 0.425 & 0.246 & 0.784 & 0.391 & 0.195 & 0.128 \\
Trajectory RAG (Summary) & Proc. & 8 & 0.408 & 0.321 & 0.822 & 0.431 & 0.184 & 0.121 \\
Template RAG (ReasonFlux) & Proc. & 8 & 0.450 & 0.342 & 0.828 & 0.417 & 0.174 & 0.210 \\
Template RAG (Self-Discover) & Proc. & 8 & 0.442 & 0.296 & 0.821 & 0.414 & 0.213 & 0.247 \\
Document RAG (CompactDS) & Fact. & 8 & 0.356 & 0.312 & 0.818 & 0.319 & 0.208 & 0.172 \\
Document RAG (Google)& Fact. & 8 & 0.483 & 0.351 & \textbf{0.838} & 0.383 & 0.246 & 0.228 \\
Length Scaling & N/A & 8 & 0.508 & 0.325 & 0.815 & 0.426 & 0.255 & 0.278 \\
Reasoning Memory & Proc. & 8 & \textbf{0.511} & \textbf{0.364} & \textbf{0.838} & \textbf{0.444} & \textbf{0.278}\dingup{67} & \textbf{0.292}\dingup{67} \\
\hdashline
Length Scaling & N/A & 30 & 0.548 & 0.358 & 0.802 & 0.447 & 0.282 & 0.302 \\
Reasoning Memory & Proc. & 30 & \textbf{0.575} & \textbf{0.392}\dingup{67} & \textbf{0.836}\dingup{67} & \textbf{0.461} & \textbf{0.310}\dingup{67} & \textbf{0.325}\dingup{67} \\
\midrule
\rowcolor{gray!20} 
\multicolumn{9}{c}{\textbf{OpenThinker3-7B}} \\
No RAG & N/A & 8 & 0.470 & 0.392 & 0.841 & 0.438 & 0.318 & 0.266 \\
Trajectory RAG (Prefix) & Proc. & 8 & 0.429 & 0.358 & 0.828 & 0.437 & 0.242 & 0.249 \\
Trajectory RAG (Summary) & Proc. & 8 & 0.438 & 0.354 & 0.855 & 0.451 & 0.242 & 0.260 \\
Template RAG (ReasonFlux) & Proc. & 8 & 0.467 & 0.354 & 0.830 & 0.457 & 0.262 & 0.268 \\
Template RAG (Self-Discover) & Proc. & 8 & 0.442 & 0.392 & 0.825 & 0.441 & 0.276 & 0.279 \\
Document RAG (CompactDS) & Fact. & 8 & 0.454 & 0.333 & 0.846 & 0.458 & 0.292 & 0.258 \\
Document RAG (Google)& Fact. & 8 & 0.501 & 0.422 & 0.864 & 0.452 & 0.235 & 0.231 \\
Length Scaling & N/A & 8 & 0.612 & 0.506 & 0.876 & 0.484 & 0.333 & 0.344 \\
Reasoning Memory & Proc. & 8 & \textbf{0.725}\dingup{67} & \textbf{0.646}\dingup{67} & \textbf{0.908}\dingup{67} & \textbf{0.522}\dingup{67} & \textbf{0.362}\dingup{67} & \textbf{0.397}\dingup{67} \\
\hdashline
Length Scaling & N/A & 30 & 0.647 & 0.528 & 0.873 & 0.502 & 0.345 & 0.343 \\
Reasoning Memory & Proc. & 30 & \textbf{0.758}\dingup{67} & \textbf{0.679}\dingup{67} & \textbf{0.911}\dingup{67} & \textbf{0.542}\dingup{67} & \textbf{0.381}\dingup{67} & \textbf{0.412}\dingup{67} \\
\midrule
\rowcolor{gray!20} 
\multicolumn{9}{c}{\textbf{Qwen3-32B}} \\
No RAG & N/A & 8 & 0.789 & 0.586 & 0.905 & 0.669 & 0.387 & 0.465 \\
Trajectory RAG (Prefix) & Proc. & 8 & 0.683 & 0.408 & 0.895 & 0.607 & 0.249 & 0.432 \\
Trajectory RAG (Summary) & Proc. & 8 & 0.629 & 0.429 & 0.895 & 0.595 & 0.267 & 0.418 \\
Template RAG (ReasonFlux) & Proc. & 8 & 0.654 & 0.500 & 0.901 & 0.644 & 0.300 & 0.278 \\
Template RAG (Self-Discover) & Proc. & 8 & 0.667 & 0.483 & 0.903 & 0.633 & 0.312 & 0.339 \\
Document RAG (CompactDS) & Fact. & 8 & 0.604 & 0.388 & 0.899 & 0.602 & 0.267 & 0.284 \\
Document RAG (Google)& Fact. & 8 & 0.795 & \textbf{0.767} & 0.919 & 0.678 & 0.382 & 0.426 \\
Length Scaling & N/A & 8 & 0.792 & 0.591 & 0.903 & \textbf{0.699} & 0.399 & 0.444 \\
Reasoning Memory & Proc. & 8 & \textbf{0.825}\dingup{67} & \textbf{0.767} & \textbf{0.926}\dingup{67} & 0.674 & \textbf{0.462}\dingup{67} & \textbf{0.503}\dingup{67} \\
\hdashline
Length Scaling & N/A & 30 & 0.812 & 0.619 & 0.908 & \textbf{0.682} & 0.405 & 0.476 \\
Reasoning Memory & Proc. & 30 & \textbf{0.838} & \textbf{0.754}\dingup{67} & \textbf{0.923}\dingup{67} & 0.681 & \textbf{0.471}\dingup{67} & \textbf{0.508}\dingup{67} \\
\bottomrule
\end{tabular}
}
\caption{\textbf{Main results of all methods in math, science, and coding domains.} The column \textit{Ret.} = type of retrieved knowledge, Proc. = procedural, Fact. = factual. Bold numbers mark the best performance under the same budget. \ding{67} = statistically significantly better than the second best via paired t-test (p $<$ 0.05). Reasoning Memory significantly outperforms baselines across LLMs and generally improves over Length Scaling across budgets.}
\label{tab:main-results}
\vspace{-4mm}
\end{table*}

\Cref{tab:main-results} summarizes the end-to-end accuracy of Reasoning Memory and all baselines. Most baselines fail to consistently outperform No RAG on frontier reasoning tasks. Averaged over all tasks, only Google-based Document RAG improves over No RAG for two models: DeepSeek-R1-Distill-Llama-8B (0.422 vs.\ 0.413) and Qwen3-32B (0.661 vs.\ 0.634). In contrast, Reasoning Memory consistently improves over No RAG by a substantial margin, reaching 0.580 versus 0.500 when averaged across all models and tasks. The gains span the evaluated 7--8B models and Qwen3-32B, with the largest improvement on OpenThinker3-7B. This shows that, under a simple RAG pipeline, reasoning models benefit much more from large-scale procedural retrieval than from web documents, full trajectories, or small template libraries.

Reasoning Memory also remains stronger than the retrieval-free Length Scaling baseline in most settings, winning 34 out of 36 comparisons across models, tasks, and the two budget settings $m \in \{8,30\}$. Increasing the inference budget further improves performance. At $m=30$, Reasoning Memory improves over No RAG by 0.12 (19.0\%) on math, 0.05 (9.8\%) on science, and 0.09 (28.9\%) on coding, averaged across models. These gains support procedural knowledge retrieval as an effective way to improve test-time scaling for reasoning models. We next examine this effect in more detail and show that the gap remains consistent as the inference budget increases and more samples are drawn. In the later sections, we also analyze the datastore composition and key design ablations in detail. Additional empty-context and majority-vote controls are reported in \S\ref{appendix-selection-controls}, and qualitative examples are provided in \S\ref{appendix-qualitative-study}.

\subsection{Scaling Behavior with Larger Inference Budgets}
\label{section:results-analysis-scaling}

\begin{figure}[t!]
% \vspace{-2mm}
\centering
\includegraphics[width=\linewidth]{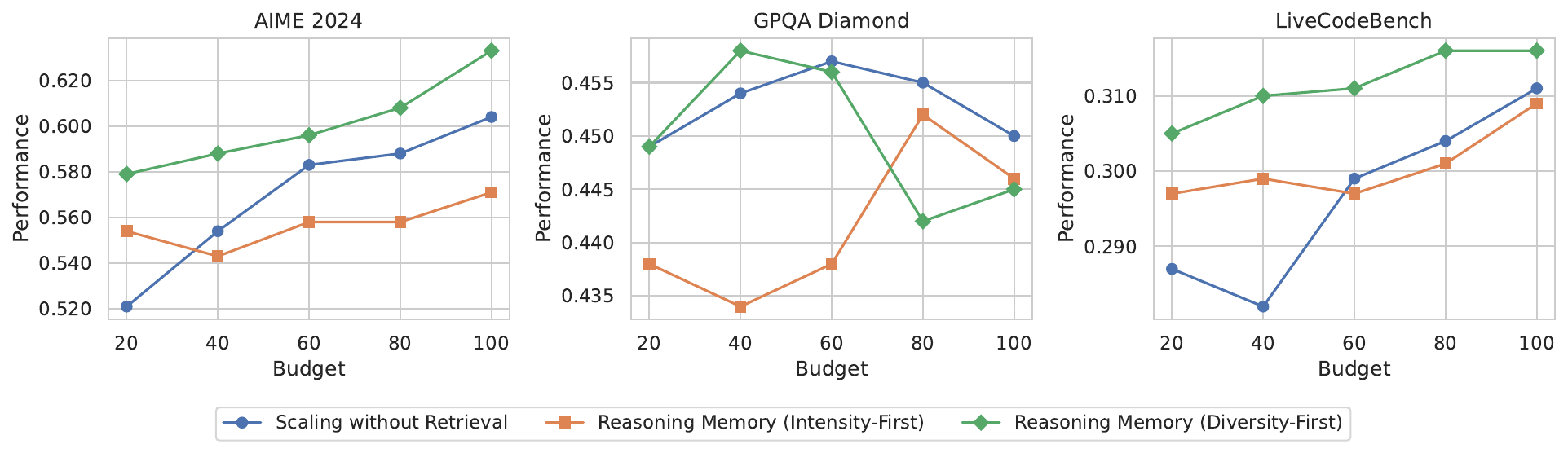}
% \vspace{-1mm}
\caption{\textbf{Performance as a function of inference budget.} We compare Length Scaling without retrieval with two Reasoning Memory variants on DeepSeek-R1-Distill-Llama-8B as the total sampling budget $m$ increases.}
\label{fig:budget-scaling}
% \vspace{-1mm}
\end{figure}

We next study whether Reasoning Memory continues to help as the inference budget increases, and how the budget should be allocated across retrieved subroutines. Using DeepSeek-R1-Distill-Llama-8B, we evaluate $m \in \{20,40,60,80,100\}$ under three strategies: (1) \emph{Length Scaling}, which samples $m$ trajectories without retrieval; (2) \emph{Reasoning Memory (Intensity-First)}, which allocates 20 samples to each retrieved subroutine and uses only a subset of the top-ranked subroutines to stay within the budget limit; and (3) \emph{Reasoning Memory (Diversity-First)}, which always uses the top 20 retrieved subroutines and allocates the maximum number of samples per subroutine within the budget limit.

As shown in \Cref{fig:budget-scaling}, the diversity-first strategy yields the strongest scaling behavior. On AIME 2024 and LiveCodeBench (V1--V4), it improves monotonically with budget and achieves the best performance at $m=100$, clearly outperforming both Length Scaling and intensity-first Reasoning Memory. On GPQA-D, all methods show weaker returns from additional budget, but diversity-first Reasoning Memory remains competitive with Length Scaling, and its best result is still achieved at a lower budget\footnote{In fact, in \Cref{tab:main-results}, the main setting reaches an even higher GPQA-D score of 0.461 with $m=30$.}. Overall, these results highlight the importance of a diversity-first budget allocation strategy and confirm the advantage of Reasoning Memory at larger budgets.

\begin{figure}[t]
% \vspace{-2mm}
    \centering
    \includegraphics[width=\linewidth]{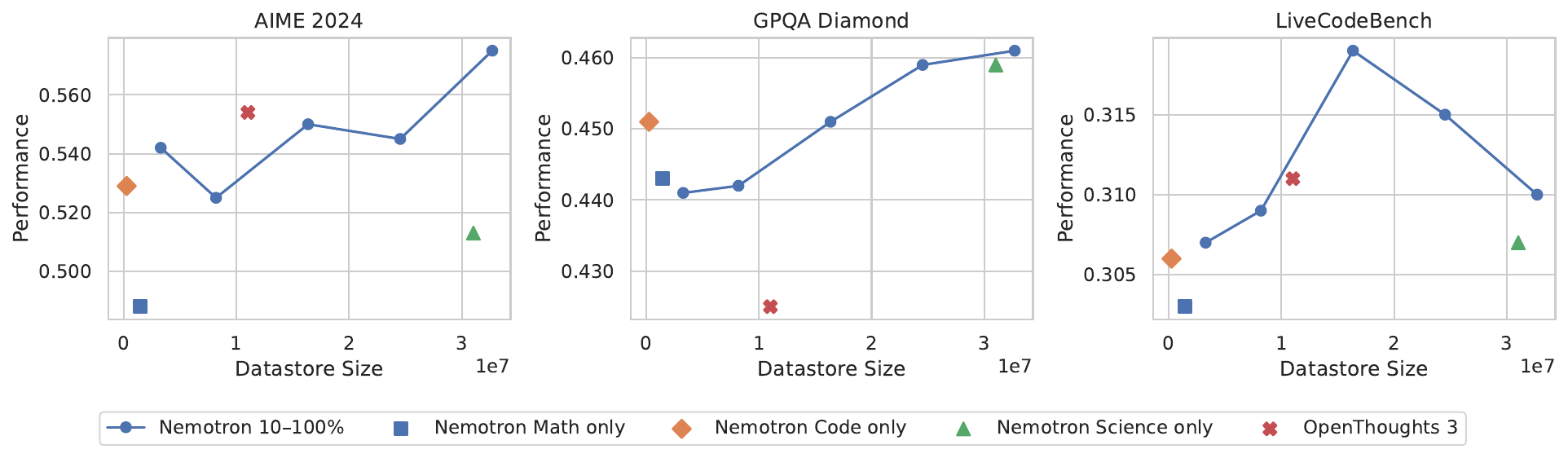}
    \vspace{-1mm}
    \caption{\textbf{Effect of datastore size and composition on Reasoning Memory.} Performance of DeepSeek-R1-Distill-Llama-8B with budget $m = 30$. Larger and more diverse datastores generally yield stronger performance.}
    \label{fig:datastore-composition-vs-performance}
    % \vspace{-2mm}
\end{figure}

\subsection{Impact of Datastore Composition}
\label{section:results-analysis-datastore}

We next study how the size and composition of the procedural datastore affect Reasoning Memory. Using DeepSeek-R1-Distill-Llama-8B with $m=30$, we compare three classes of datastore variants: (1) random subsets of Nemotron ranging from 10\% to 100\% of trajectories, (2) domain-specific Nemotron subsets containing only math, code, or science trajectories, and (3) a datastore built from OpenThoughts3, which is more math-focused.

\Cref{fig:datastore-composition-vs-performance} shows that performance generally improves as we scale the mixed Nemotron datastore. The full datastore gives the best accuracy on AIME 2024 and GPQA-D, while LiveCodeBench is less sensitive once the datastore reaches roughly 25\% of full scale. Among domain-specific datastores, OpenThoughts3 is highly competitive on AIME 2024, matches similarly sized Nemotron subsets on LiveCodeBench, but underperforms clearly on GPQA-D. Within Nemotron, science-only trajectories give the best science performance on GPQA-D, while code-only trajectories are surprisingly more effective than math-only trajectories across all three tasks despite having similar size. Code trajectories may transfer through domain-general patterns such as decomposition, constraint tracking, case analysis, debugging, and verification. For both math and coding, however, all in-domain subsets underperform the full mixed datastore, suggesting that broader procedural coverage is useful. These results do not identify which entries, or what fraction of the full datastore, drive the gains.

\begin{table}[t]
\centering
\resizebox{\linewidth}{!}{
\begin{tabular}{lcccc}
\toprule
Setting & AIME 2024 & GPQA-D & LCB (V1-4) & Average \\
\midrule
Removing Subroutine Decomposition & 0.551 & 0.433 & 0.308 & 0.431 \\
Removing Self-Generated Query & 0.567 & 0.453 & 0.295 & 0.438 \\
Full Reasoning Memory (Qwen3-8B) & 0.550 & \textbf{0.474} & \textbf{0.320} & 0.448 \\
Full Reasoning Memory (QwQ-32B) & \textbf{0.575} & 0.461 & 0.310 & \textbf{0.449} \\
\bottomrule
\end{tabular}
}
\caption{\textbf{Ablations on datastore construction and query design.} All results use DeepSeek-R1-Distill-Llama-8B with budget $m = 30$. The full Reasoning Memory configuration uses QwQ-32B as the datastore generator, a decomposed datastore indexed by subquestions, and self-verbalized in-thought queries.}
% \vspace{-4mm}
\label{tab:ablations-other}
\end{table}

\subsection{Ablations: Subroutine Decomposition, Query Generation, and Alternative Models}
\label{section:results-analysis-ablations}

We ablate three key design choices of Reasoning Memory using DeepSeek-R1-Distill-Llama-8B with $m=30$. First, we remove the subquestion decomposition and instead index the datastore by the original question, with a single subroutine summarizing the full trajectory. Second, we keep the decomposed datastore but disable self-verbalized retrieval queries, using the original question directly as the query. Third, we replace QwQ-32B with the smaller Qwen3-8B for datastore generation. Finally, the uncertainty criterion is another important design choice, which we study in \S\ref{appendix-uncertainty-metrics} due to space constraints.

As shown in \Cref{tab:ablations-other}, Reasoning Memory is robust to the choice of the datastore generator model, but both decomposition and self-verbalized queries matter. Using Qwen3-8B yields performance close to the full system, and even slightly improves GPQA-D and LiveCodeBench, suggesting that much of the useful procedural knowledge already resides in the source trajectories. In contrast, removing subroutine decomposition causes the largest drop on AIME and GPQA-D, indicating that fine-grained decomposition is important for building a diverse datastore that matches different subquestion granularities. Replacing self-verbalized queries with the original question also leads to consistent degradation across all three benchmarks. The largest drop is seen on LiveCodeBench, where problem statements are often long and cluttered with context, format specifications, and examples. Since AIME 2024 contains only 30 problems, small single-task differences should not be overinterpreted; our component conclusions rely on the combined pattern across AIME 2024, GPQA-D, and LiveCodeBench. The stronger end-to-end evidence in \Cref{tab:main-results} spans three models, six benchmarks, and two sampling budgets, with significance markers reported. Overall, these ablations support the core design choices of Reasoning Memory while suggesting that datastore construction can be performed with smaller models at little cost.

Taken together, our experiments examine four main design axes of a RAG-style memory system: the retrieved value, through documents, trajectories, summaries, templates, and subroutines (\Cref{tab:main-results}); the key and query, through full-question indexing and original-question queries (\Cref{tab:ablations-other}); and utilization and selection, through budget allocation (\Cref{fig:budget-scaling}) and alternative uncertainty criteria (\Cref{tab:appendix-alt-metrics}).

\section{Conclusion}
\label{section:conclusion}

Reasoning Memory shows that test-time compute for reasoning models can be spent not only on producing longer chains of thought, but also on reusing prior problem-solving experience in a structured way. By surfacing procedural knowledge from existing reasoning trajectories and turning it into an explicit datastore, our approach transforms instruction-tuning corpora into an inference-time resource for guiding new problem-solving episodes. We further introduce a simple recipe for aligning retrieval with the model’s own thinking stream and for scaling inference with multiple retrieved priors. Empirically, Reasoning Memory improves diverse open-weight reasoning models on math, science, and coding tasks, scales well with both datastore size and inference budget, and remains robust across several implementation choices. More broadly, our results suggest a promising direction for both the reasoning and RAG communities: building systems that can accumulate, retrieve, and reuse procedural knowledge over time.

\section*{Ethics Statement}

\paragraph{Data and Reproducibility}
All datasets used in this work are existing public  resources and are used as-is for evaluation and corpus construction. We do not redistribute any model outputs, datasets, or source corpora. Our experiment code and scripts for reproducing the corpus construction pipeline from the original sources are publicly available at \url{https://github.com/facebookresearch/reasoning-memory}. Corpus statistics are reported in \S\ref{section-approach-datastore}. This work does not involve human annotators or human subjects.

\paragraph{LLM Use in Research}
LLMs were used in three limited ways during this project. First, all figures and tables, except \Cref{fig:main-figure}, were produced with LLM assistance using statistics and data provided by the authors. The authors fully verified the correctness of these materials and take full responsibility for them. Second, LLMs and coding agents, specifically OpenAI ChatGPT and OpenAI Codex, were used to assist with implementing data processing scripts, and the authors fully verified the correctness of the resulting code. Third, LLMs, specifically OpenAI ChatGPT and Google Gemini, were used to help polish the writing of the paper from a fully human-written draft.

\paragraph{Limitations}
Our datastore is built from reasoning trajectories that span the evaluated math, science, and coding domains. Although Nemotron applies an explicit test-contamination removal process \citep{NemotronPostTrainingDatasetV1}, this alignment does not eliminate benchmark-overfitting risk. Our domain-specific and OpenThoughts3 results provide evidence of cross-domain and cross-source transfer, but do not establish generalization to arbitrary out-of-domain procedural corpora. We also do not measure what fraction of the 32M entries drives the gains or evaluate reasoning models below 7B parameters.

\bibliography{colm2026_conference}
\bibliographystyle{colm2026_conference}

\clearpage
\newpage
\appendix

\section{Additional Pilot Study Details}
\label{appendix-pilot-study}

\subsection{Standard Document RAG with CompactDS}
\label{appendix-pilot-compactds}

As a pilot study, we investigate how a standard document-level RAG pipeline affects \emph{instruction-tuned} models versus \emph{reasoning} models. We follow the CompactDS pipeline of \citet{DBLP:journals/corr/abs-2507-01297}, which retrieves general-domain background passages and appends them to the model prompt. For each benchmark question, we use the question text as the retrieval query, perform the same two-stage retrieval procedure as the original implementation, and insert the top-$k$ retrieved documents into the prompt using the same formatting and instruction template. Unless otherwise stated, we set $k=3$.

We evaluate paired instruction-tuned and reasoning models from three model families: Llama-3.1-8B \citep{llama3}, Qwen2.5-7B, and Qwen2.5-32B \citep{qwen2_5}. For Llama-3.1-8B and Qwen2.5-7B, we use DeepSeek-R1 distillations as the reasoning variants \citep{dsr1}. For Qwen2.5-32B, we use QwQ-32B as the reasoning variant \citep{qwq}. We evaluate on AIME 2024 \citep{aime2024}, GPQA-Diamond \citep{gpqa}, and LiveCodeBench \citep{livecodebench}. For each question, we generate $m=8$ independent samples with a maximum output length of 32k tokens and report the average performance across samples. Following \citet{DBLP:journals/corr/abs-2501-05366}, we use math-equal for AIME, exact match for GPQA-D, and execution-based pass@1 for LiveCodeBench.

\Cref{fig:compactds-instr-reason} shows a consistent pattern across model families and tasks. Instruction-tuned models operate in a lower-accuracy regime but obtain modest and fairly reliable gains from retrieved documents, consistent with the findings of \citet{DBLP:journals/corr/abs-2507-01297}. In contrast, the corresponding reasoning models start from substantially stronger no-retrieval performance, yet standard document RAG provides only limited gains and can reduce accuracy in multiple settings. The negative effect is most visible for smaller reasoning models, but the overall pattern is consistent across math, science, and coding tasks.

We view this result as evidence of a mismatch between standard document RAG and reasoning models. First, reasoning models are not explicitly trained to treat retrieved passages as procedural guidance that should steer an ongoing chain of thought. Second, retrieved web documents are optimized for broad knowledge coverage rather than for the specific subquestion currently being solved. As a result, even relevant retrievals may be only weakly aligned with the model's active reasoning state. This motivates the central design choice of our method: instead of retrieving general documents, we retrieve compact procedural subroutines that more directly match the current subquestion.

\subsection{Controlled Knowledge Injection: Factual vs. Procedural}
\label{appendix-pilot-knowledge}

The CompactDS result above leaves open an important question: are reasoning models inherently incompatible with retrieval, or is the issue the form of the retrieved knowledge? To probe this question, we run a controlled knowledge injection study in which the retrieved content is synthesized to be directly relevant to each evaluation problem.

For each problem from AIME 2024, GPQA-Diamond, and LiveCodeBench, we prompt a strong model (GPT-5.2 via API, \citet{openai_gpt52_2025}) to synthesize a short paragraph that is helpful for solving the problem while satisfying two constraints: (1) it must not reveal the final answer, and (2) it must not provide intermediate computations or step-by-step derivations that could leak the answer. We synthesize two types of content: (1) \textit{factual knowledge}, including definitions, formulas, and theorems, and (2) \textit{procedural knowledge}, including high-level decomposition strategies, solution plans, and verification heuristics. The prompts are shown in \Cref{fig:prompt-knowledge-synthesis-factual,fig:prompt-knowledge-synthesis-procedural}. The resulting paragraph is typically 100--200 tokens long and is inserted into the same prompt location used for CompactDS documents.

\begin{figure}[t]
\centering
\includegraphics[width=0.9\linewidth]{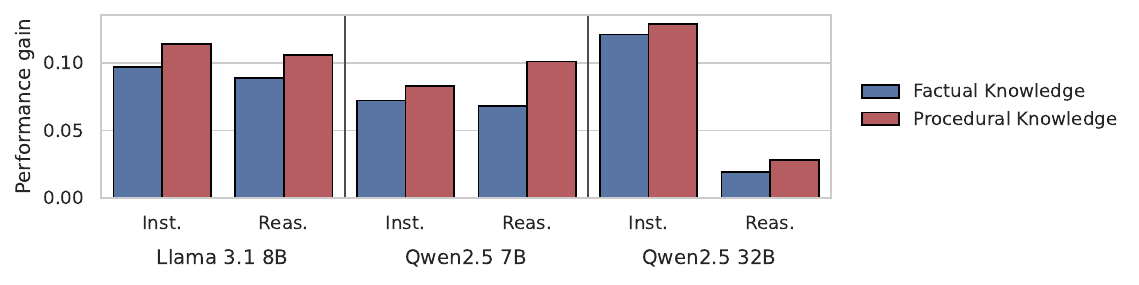}
\caption{\textbf{Utility of different types of synthesized knowledge.} We report performance gains relative to the no-retrieval setting. Both factual and procedural knowledge help, but procedural knowledge yields larger gains across tasks and model families.}
\label{fig:pilot-fact-proc}
\end{figure}

\Cref{fig:pilot-fact-proc} reports performance deltas relative to the no-retrieval setting. Both instruction-tuned and reasoning models improve when given synthesized knowledge, which suggests that retrieval itself is not the core issue. Instead, the usefulness of retrieval depends strongly on the form and alignment of the injected content. In particular, reasoning models benefit much more from these targeted knowledge paragraphs than from retrieved CompactDS documents, even though both are inserted through the same prompt template. This supports the view that reasoning models are sensitive to the relevance and specificity of the injected context.

Procedural knowledge is consistently more helpful than factual knowledge across tasks and model families. We hypothesize that this is because many difficult reasoning problems are bottlenecked less by missing background facts than by missing guidance on how to decompose the problem, choose a strategy, or verify an intermediate direction. This pattern is especially natural for AIME and LiveCodeBench, which often require problem reframing and multi-step inference. GPQA-Diamond benefits somewhat more from factual knowledge, likely because some questions can be answered more directly from domain concepts or definitions, but procedural guidance remains stronger overall.

These results motivate our focus on building a scalable datastore that explicitly captures procedural knowledge and retrieves it in a form that reasoning models can readily use. \Cref{tab:pilot-fact-vs-proc-full-results} reports the full per-task results for factual and procedural knowledge injection. The “Average” row corresponds to the aggregate numbers shown in \Cref{fig:pilot-fact-proc}.

\begin{table}[t]
\centering
\small
\setlength{\tabcolsep}{5pt}
\resizebox{\linewidth}{!}{
\begin{tabular}{lcccccccccccc}
\toprule
& \multicolumn{2}{c}{Llama-3.1-8B (Inst.)}
& \multicolumn{2}{c}{Llama-3.1-8B (Reas.)}
& \multicolumn{2}{c}{Qwen2.5-7B (Inst.)}
& \multicolumn{2}{c}{Qwen2.5-7B (Reas.)}
& \multicolumn{2}{c}{Qwen2.5-32B (Inst.)}
& \multicolumn{2}{c}{Qwen2.5-32B (Reas.)} \\
\cmidrule(lr){2-3}\cmidrule(lr){4-5}\cmidrule(lr){6-7}\cmidrule(lr){8-9}\cmidrule(lr){10-11}\cmidrule(lr){12-13}
& Fact. & Proc.
& Fact. & Proc.
& Fact. & Proc.
& Fact. & Proc.
& Fact. & Proc.
& Fact. & Proc. \\
\midrule
AIME & 0.040 & 0.029 & 0.075 & 0.090 & 0.015 & 0.012 & 0.096 & 0.167 & 0.025 & 0.067 & -0.013 & -0.023 \\
GPQA & 0.223 & 0.254 & 0.150 & 0.151 & 0.173 & 0.204 & 0.076 & 0.084 & 0.159 & 0.160 & 0.065  & 0.076  \\
LCB & 0.027 & 0.059 & 0.041 & 0.077 & 0.029 & 0.033 & 0.033 & 0.053 & 0.178 & 0.159 & 0.004  & 0.032  \\
\midrule
Average  & 0.097 & 0.114 & 0.089 & 0.106 & 0.072 & 0.083 & 0.068 & 0.101 & 0.121 & 0.129 & 0.019  & 0.028  \\
\bottomrule
\end{tabular}
}
\caption{\textbf{Per-task performance gains in the knowledge-synthesis pilot study.} ``Fact.'' denotes factual knowledge and ``Proc.'' denotes procedural knowledge. Entries report the performance difference between prompting with synthesized knowledge and prompting without augmentation. Procedural knowledge generally produces larger gains.}
\label{tab:pilot-fact-vs-proc-full-results}
\end{table}

\begin{figure}[h!]
\begin{tcolorbox}[colback=chocolate!5!white,colframe=chocolate!75!black,
                  title=Prompt for Synthesizing Factual Knowledge]
\begin{VerbatimWrap}
You will read a question and write one short passage of factual background knowledge. Directly write general definitions, concepts, and relationships that might be relevant. Do not state the connection between the knowledge and the question. Do not include any steps toward solving the question.

Question:
{question}

Write the paragraph now.
\end{VerbatimWrap}
\end{tcolorbox}
\caption{\textbf{Prompt for synthesizing factual knowledge for the pilot study.} GPT-5.2 is prompted to generate the knowledge.}
\label{fig:prompt-knowledge-synthesis-factual}
\end{figure}

\begin{figure}[h!]
\begin{tcolorbox}[colback=chocolate!5!white,colframe=chocolate!75!black,
                  title=Prompt for Synthesizing Procedural Knowledge]
\begin{VerbatimWrap}
You will read a question and write one short paragraph of procedural knowledge (generalizable steps/strategies) as the hint to solving the problem. Describe how someone should approach this general type of problem step-by-step. Keep it high-level and reusable: do NOT compute anything, do NOT include concrete numbers or intermediate results.

Question:
{question}

Write the paragraph now.
\end{VerbatimWrap}
\end{tcolorbox}
\caption{\textbf{Prompt for synthesizing procedural knowledge for the pilot study.} GPT-5.2 is prompted to generate the knowledge.}
\label{fig:prompt-knowledge-synthesis-procedural}
\end{figure}

\begin{figure}[ht]
\begin{tcolorbox}[colback=chocolate!5!white,colframe=chocolate!75!black,,title=Prompt for Trajectory Decomposition]
\begin{VerbatimWrap}
You are an expert tutor. Given a question, a final answer, and a detailed solution process written by the teacher, break the problem down into a list of general, self-contained subquestions (SQs). Each subquestion should capture a logical reasoning step or reusable subroutine that can help solve similar problems. The subquestions should be general and you should avoid using the original question's surface form. Avoid introducing specific numbers or setups as much as you can. In your response, start your subquestions under a section ### Subquestions .
Write one question on each line, starting by an index 1. 2. 3., etc.

### Question:
{question}

### Teacher Answer:
{teacher_answer}

### Teacher Thought Process:
{teacher_thought}

Now, analyze the provided information and list the important subquestions.
\end{VerbatimWrap}
\end{tcolorbox}
\caption{\textbf{Prompt for decomposing trajectory into the subquestions.} }
\label{fig:prompt-subq-decomp}
\end{figure}

\begin{figure}[ht]
\begin{tcolorbox}[colback=chocolate!5!white,colframe=chocolate!75!black,title=Prompt for Subroutine Generation over Subquestions]
\begin{VerbatimWrap}
You are an expert tutor. Given a question, a final answer, a detailed solution process, and a subquestion (SQ),
write a reusable subroutine-style hint that describes how to solve SQ.
The hint should restate the general subproblem it addresses, and then describe the reasoning steps one might follow.
Use a first-person voice like "For problems like this, I should...".
In your response, start by outlining the general problem under a section ### Applicable Problems .
Then, start your hint on a new line by ### Hint.
Inside the hint, you must first re-state the general problem setting that the hint can apply to.

### Question:
{q}

### Teacher Answer:
{ans}

### Teacher Thought Process:
{thought}

### Subquestion:
{subq}


Now, analyze the question, answer, and teacher's thought and write your hint for the subquestion.
Make sure your hint helps approach similar questions without revealing the answer or any intermediate results.
\end{VerbatimWrap}
\end{tcolorbox}
\caption{\textbf{Prompt for generating reusable subroutines for subquestions.}}
\label{fig:prompt-subq-hint}
\end{figure}

\section{Additional Implementation Details}
\label{appendix-impl-details}

\subsection{Reasoning Memory}
\label{appendix-reasoning-memory-impl-details}

\begin{figure}[h!]
\begin{tcolorbox}[colback=chocolate!5!white,colframe=chocolate!75!black]
\begin{VerbatimWrap}
\end{VerbatimWrap}
\small
\underline{\textbf{Original Question:}}

A museum prints ticket codes in an unknown base $b\ge 6$. One ticket reads $2A5_b$, where $A$ is a single digit. When interpreted as a base-10 integer, the code must (1) be the product of two distinct primes and (2) satisfy the checksum $N\equiv 3 \pmod{11}$. Determine all pairs $(b,A)$ that produce a valid ticket.\\

\underline{\textbf{Original Reasoning Trajectory:}}

\texttt{Okay, so I've got this math problem here... }(very long, omitted)
\\

\underline{\textbf{Subquestions:}}

[SQ1] How do you verify that a given numeral is valid in base $b$?

[SQ2] How do you convert a number from an arbitrary base $b$ to its decimal (base 10) equivalent?

[SQ3] How do you determine whether a number is a product of two distinct primes?

[SQ4] How do you solve a modular arithmetic constraint of the form $N \equiv r \pmod{m}$ when N is an algebraic expression in unknown integers?
\\

\underline{\textbf{Subroutine for SQ2:}}

For problems like this, I should first write the number as a sum of each digit multiplied by the base raised to the power of its position index. Here's how:
    \begin{enumerate}
        \item \textbf{Identify each digit} of the number in the original base.
        \item \textbf{Assign positional indices} starting from $0$ at the rightmost digit (the least significant digit).
        \item \textbf{Compute the weight} of each digit by raising the base $b$ to the power of its position index.
        \item \textbf{Multiply each digit} by its corresponding weight.
        \item \textbf{Sum all the products} to obtain the decimal (base 10) equivalent.
    \end{enumerate}
    For example, a number $d_n d_{n-1} \ldots d_1 d_0$ in base $b$ converts to decimal as:
    \[
        d_n \cdot b^n + d_{n-1} \cdot b^{n-1} + \ldots + d_1 \cdot b^1 + d_0 \cdot b^0.
    \]
    Ensure all digits are valid in the original base (i.e., $0 \le d_i < b$) before proceeding.

\end{tcolorbox}
\caption{\textbf{An Example of the Reasoning Memory Datastore.} We extract multiple self-contained subquestions from the original question and the long reasoning trajectory. Then, a corresponding subroutine is generated for each subquestion.}
\label{fig:example-datastore-content}
\end{figure}

\paragraph{Datastore Construction} We leverage the publicly released Nemotron post-training V1 corpus\footnote{Downloaded at \url{https://huggingface.co/datasets/nvidia/Nemotron-Post-Training-Dataset-v1}.}. We use the math, stem, and code subsets, which contain 2.0, 20.7, and 1.9 million trajectories, respectively. Nemotron pairs each question with multiple reasoning trajectories generated with a teacher model. Empirically, we found that these trajectories often converge to similar general approaches and differ mostly in local steps or formatting. As such, decomposing a large number of trajectories for each question would bring marginal gains in datastore diversity. We thus select the first trajectory per question to construct the datastore. After this deduplication process, we obtained roughly 170k unique questions for math, 3 million for stem, and 34k for code, comprising approximately 10\% of the original dataset. Each trajectory contains a user question and a response generated by DeepSeek-R1. We prompt QwQ-32B with the prompt shown in \Cref{fig:prompt-subq-decomp} to generate subquestions from the full trajectories with temperature=0.6, top\_p=0.95, and max\_tokens=10,000. Next, we use the prompt shown in \Cref{fig:prompt-subq-hint} to generate a subquestion-specific subroutine conditioned on the original question, trajectory, and subquestion. For this step, we prompt QwQ-32B with the same parameters. The final datastore contains 32 million (subquestion, subroutine) pairs. On average, we obtain 10.5 subquestions per trajectory. Each subquestion contains 17.7 words on average (19.2 tokens with the QwQ-32B tokenizer), and each subroutine contains 197.6 words on average (207.9 tokens), significantly fewer than the original trajectories.

\paragraph{Retrieval} We leverage ReasonIR-8B for all the reasoning memory experiments in the paper\footnote{Accessed at \url{https://huggingface.co/reasonir/ReasonIR-8B}}. The subquestions are encoded and used as the keys. We do not add special instructions but only added the special token \texttt{"<|embed|>"} as used in the official codebase. For the queries, we prepend an additional domain-specific instruction before encoding with ReasonIR, as shown below: 

\begin{center}
    (For Math) \texttt{Please answer the following math question.} \\
    (For Coding) \texttt{Generate a correct Python program that passes \\ all tests for the given problem.} \\
    (For Science) \texttt{Please answer the following question.} \\
\end{center}

\paragraph{RAG Inference} We perform query verbalization and final generation using the prompt presented in \S\ref{section-approach-retrieval}. For query verbalization, we add the thought hijacking prompt immediately after the question and generate a maximum of 100 tokens. Then, we use the nltk toolkit \citep{bird-loper-2004-nltk} to segment the first sentence in the model's response as the query. For the final answer generation, we use top-p sampling with temperature 0.7 and $p=0.95$ for Qwen3-32B and $p=0.8$ for the other two small models. The maximum number of output tokens is set to 32,768. This same set of parameters is used for all the baselines as well. All the experiments are run with a supercomputing cluster containing A100, H100, and H200 nodes. We host the models via vllm \citep{vllm} and perform batched calls. 

\paragraph{Sample Filtering} We perform additional filtering over the $m$ samples collected from top-$k$ retrievals. For $\pi_{j,l}$, sample $l$ from the $j^{th}$ retrieved hint, we calculate the raw score $r_{j,l}$ as $|\pi_{j,l}|$, the trajectory length in tokens. Let $r_{max}$ and $r_{min}$ be the maximum and minimum scores across all samples for the same question. We normalize each score as $\hat{r}_{j,l}=(r_{max}-r_{j,l})/(r_{max}-r_{min})\in[0, 1]$. Since shorter responses are preferred, a higher $\hat{r}_{j,l}$ is better. To filter retrieved subroutines, we calculate $\hat{r}_j$ by averaging the samples corresponding to each subroutine and retain those with scores higher than $\tau=0.1$. Finally, we sort the samples corresponding to the retained subroutines and retain the $n$ samples with the highest $\hat{r}_{j,l}$. If fewer than $n$ samples remain, we backfill from the filtered-out pool in descending order of $\hat{r}_{j,l}$. Performance is locally stable for $\tau \in [0.1,0.3]$. With DeepSeek-R1-Distill-Llama-8B, an average of 2.6 and 5.3 subroutines remain for $(m,n)=(8,4)$ and $(30,8)$, respectively, averaged across datasets; the fallback is not invoked at $\tau=0.1$ in these evaluations.

\paragraph{Efficiency} The online retrieval overhead is small relative to long-form generation. On an A100 with Qwen3-8B, query generation takes less than one second, and retrieval and subroutine injection together take less than one second, compared with approximately 10 seconds for one long reasoning sample. With batched generation, the $m=30$ setting takes approximately 60 seconds end-to-end. Reasoning Memory and Length Scaling use the same number of generated trajectories, while the $m=100$ experiments are intended only to study scaling behavior. Datastore construction is a one-time offline cost. By entry count, our 32M-entry datastore is also substantially smaller than CompactDS, which contains 1.85B 256-word chunks \citep{DBLP:journals/corr/abs-2507-01297}, and Google search operates over a web-scale corpus.

\subsection{Baselines}
\label{appendix-baseline-impl-details}

In this section, we outline the implementation of the baseline methods. 

\begin{itemize}
    \item \textbf{Trajectory RAG (Prefix)} We index the deduplicated Nemotron trajectories using their questions as the key and ReasonIR as the embedding model. The test questions are used as the query. We use the first 256 tokens of the top-3 retrieved trajectories as we empirically found it to cover the first paragraph, which is often a high-level plan of the problem solving strategy. We format the retrieved trajectories in the same way as Reasoning Memory. 
    \item \textbf{Trajectory RAG (Summary)} We use the same retrieval methods as Trajectory RAG (Prefix). The only difference is that we prompt the QwQ-32B  to extract the high-level question from the trajectory and write a corresponding subroutine using a prompt similar to \Cref{fig:prompt-subq-decomp} and \Cref{fig:prompt-subq-hint}. We use top-1 retrieved trajectory as it empirically leads to better performance than top-3 or top-5.  
    \item \textbf{Template RAG (ReasonFlux)} We access the templates from ReasonFlux's official repository \footnote{\url{https://github.com/Gen-Verse/ReasonFlux/blob/d82429dae582c8ecbe762d643c909955e7cb9636/ReasonFlux_v1/data/template_library.json}}. After preprocessing, we collected a corpus with 420 solution templates for various math problems. We concatenate the template metadata and the content as the key for retrieval. Using ReasonIR, we retrieve top-1 relevant template and evaluate using the same prompt as Reasoning Memory. 
    \item Template RAG (Self-Discover) We convert the templates directly from the appendix section in \citep{DBLP:conf/nips/ZhouP0CCLCZMZ24} and ended up with a corpus containing 39 high-level problem solving strategies. We concatenate the template metadata and the content as the key for retrieval. Using ReasonIR, we retrieve top-1 relevant template and evaluate using the same prompt as Reasoning Memory.
    \item \textbf{Document RAG (CompactDS)} We use the same implementations for the pilot study. Using the official codebase and data\footnote{\url{https://github.com/Alrope123/compactds-retrieval}}, we retrieve and augment top-3 documents via two-step exact search. 
    \item \textbf{Document RAG (Google)} We use the baseline implementation provided by \citet{DBLP:journals/corr/abs-2504-21776}. We use Serper search and augment the top-10 retrieved document in the context following the original implementation\footnote{\url{https://github.com/RUC-NLPIR/WebThinker/blob/main/scripts/run_naive_rag.py}}. In our preliminary studies, we found the default Serper search results contain a large number of data contamination instances, where the webpage content directly contains the answer. We therefore set the date cutoff to 2023/12/31, which we found to minimize the contaminations. 
    \item \textbf{Length Scaling} We use the same implementation as in the pilot study to repeatedly generate samples. Our implementation was inspired by \citet{DBLP:journals/corr/abs-2501-05366}'s baseline implementation\footnote{\url{https://github.com/RUC-NLPIR/Search-o1/blob/main/scripts/run_direct_gen.py}}. 
\end{itemize}

\section{Additional Analyses}

\subsection{Empty-Context and Majority-Vote Controls}
\label{appendix-selection-controls}

To isolate the contribution of retrieval from query verbalization, we replace the retrieved subroutine with an empty context while retaining query verbalization and length-based selection. We also compare against majority voting on AIME 2024 and GPQA-D, where answer aggregation is well defined. As shown in \Cref{tab:empty-context-majority-vote}, Reasoning Memory is strongest in three of the four model-task comparisons, while majority voting is stronger for DeepSeek-R1-Distill-Llama-8B on GPQA-D. Empty context is consistently weaker than Reasoning Memory on AIME and does not explain its overall gains.

\begin{table}[h]
\centering
\resizebox{0.8\linewidth}{!}{
\begin{tabular}{llcc}
\toprule
Model & Method & AIME 2024 & GPQA-D \\
\midrule
OpenThinker3-7B
& Length Scaling & 0.647 & 0.502 \\
& Empty Context + Length Scaling & 0.547 & 0.511 \\
& Majority Vote & 0.573 & 0.533 \\
& Reasoning Memory & \textbf{0.758} & \textbf{0.542} \\
\midrule
DeepSeek-R1-Distill-Llama-8B
& Length Scaling & 0.548 & 0.447 \\
& Empty Context + Length Scaling & 0.555 & 0.452 \\
& Majority Vote & 0.496 & \textbf{0.472} \\
& Reasoning Memory & \textbf{0.575} & 0.461 \\
\bottomrule
\end{tabular}
}
\caption{\textbf{Empty-context and majority-vote controls.} All results use $m=30$. Empty Context retains query verbalization but provides no retrieved subroutine before applying length-based selection. Majority voting is reported only for AIME 2024 and GPQA-D, where answer aggregation is well defined.}
\label{tab:empty-context-majority-vote}
\end{table}

\subsection{Combining Baselines with Length Scaling}
\label{appendix-baselines-with-len-scaling}

In the main text, we only apply length-based uncertainty filtering to the retrieval-free Length Scaling baseline and to Reasoning Memory. For completeness, \Cref{tab:appendix-length-rag} reports results when we additionally apply the same filtering procedure to other RAG baselines under the $m=8$ budget, further reducing the pool to $n=4$ samples. Length filtering generally improves performance for the smaller reasoning models, especially for No RAG and Document RAG. For the larger Qwen3-32B model, however, the effect is mixed: length filtering helps No RAG and Document RAG but consistently hurts Trajectory RAG and Template RAG. Since this step is not a standard practice in prior work and its impact is not uniform across settings, we do not include these variants in the main comparisons.

\begin{table}[h]
\centering
\resizebox{\linewidth}{!}{
\begin{tabular}{lccc|ccc|ccc}
\toprule
& \multicolumn{3}{c|}{\textbf{DeepSeek-R1-Distill-Llama-8B}}
& \multicolumn{3}{c|}{\textbf{OpenThinker3-7B}}
& \multicolumn{3}{c}{\textbf{Qwen3-32B}} \\
\cmidrule(lr){2-4} \cmidrule(lr){5-7} \cmidrule(lr){8-10}
\textbf{Method} & \textbf{AIME 24’} & \textbf{GPQA} & \textbf{LCB (V1–4)} & \textbf{AIME 24’} & \textbf{GPQA} & \textbf{LCB (V1–4)} & \textbf{AIME 24’} & \textbf{GPQA} & \textbf{LCB (V1–4)} \\
\midrule
No RAG & 0.461 & 0.427 & 0.220 & 0.470 & 0.438 & 0.318 & 0.789 & 0.669 & 0.387 \\
\quad + length scaling & 0.508 & 0.426 & 0.255 & 0.612 & 0.484 & 0.333 & 0.792 & 0.699 & 0.399 \\
Trajectory RAG (Summary)     & 0.408 & 0.431 & 0.184 & 0.438 & 0.451 & 0.242 & 0.629 & 0.595 & 0.267 \\
\quad + length scaling & 0.483 & 0.439 & 0.175 & 0.533 & 0.444 & 0.290 & 0.567 & 0.580 & 0.224 \\
Template RAG (Self-Discover) & 0.442 & 0.414 & 0.213 & 0.442 & 0.441 & 0.276 & 0.667 & 0.633 & 0.312 \\
\quad + length scaling & 0.500 & 0.431 & 0.252 & 0.533 & 0.451 & 0.303 & 0.608 & 0.623 & 0.215 \\
Document RAG (Google) & 0.483 & 0.383 & 0.246 & 0.501 & 0.452 & 0.235 & 0.795 & 0.678 & 0.382 \\
\quad + length scaling & \textbf{0.523} & 0.390 & 0.265 & 0.475 & 0.432 & 0.250 & 0.815 & \textbf{0.682} & 0.436 \\
Reasoning Memory       & 0.511 & \textbf{0.444} & \textbf{0.278} & \textbf{0.725} & \textbf{0.522} & \textbf{0.362} & \textbf{0.825} & 0.674 & \textbf{0.462} \\
\bottomrule
\end{tabular}
}
\caption{\textbf{Effect of adding length-based uncertainty filtering to retrieval-based baselines.} All results use $m=8$ and $n=4$.}
\label{tab:appendix-length-rag}
\end{table}

\subsection{Retrieval vs. Training}
\label{appendix-sft}

% \diwu{Mingda: emphasize that this is not an entirely apple-to-apple comparison}

Reasoning Memory provides a model-agnostic way to digest a reasoning corpus as an external datastore and perform RAG at inference time. Alternatively, post-training can internalize the corpus into a model's parameters. Is Reasoning Memory still useful after the same corpus has been internalized? We investigate this question using the Llama-3.1-Nemotron-Nano-8B-v1 model released by Nvidia. According to the model card, this model was likely trained on variants of the Nemotron v1 corpus via supervised fine-tuning (SFT) and reinforcement learning (RL). We emphasize that this comparison is not fully controlled for two reasons: (1) the detailed training data composition and compute recipe are not entirely clear, and (2) due to question deduplication, we use a Nemotron subset approximately 10\% as large as the original corpus. A fully controlled comparison would re-run post-training from the backbone model, which is beyond our budget.

As shown in \Cref{tab:reasoning-memory-vs-sft}, we observe mixed results when comparing Reasoning Memory with No RAG and Length Scaling. Although the model was trained on a similar corpus, Reasoning Memory still improves over No RAG on AIME and LCB. However, it does not outperform Length Scaling on GPQA or LCB. These results indicate that fine-tuning is a strong way to learn from a corpus and that the contribution of Reasoning Memory may depend on how much of the corpus the model has already internalized. This also opens several future directions, such as expanding the knowledge sources, actively learning the memory corpus, and learning subroutines that complement the model's existing abilities.

\begin{table}[h]
\centering
\resizebox{0.5\linewidth}{!}{
\begin{tabular}{lccc}
\toprule
\textbf{Method} & \textbf{AIME 24’} & \textbf{GPQA} & \textbf{LCB (V1–4)} \\
\midrule
No RAG & 0.658 & \textbf{0.553} & 0.380 \\
Length Scaling  & 0.692 & 0.548 & \textbf{0.396} \\
Reasoning Memory& \textbf{0.708} & 0.534 & 0.395 \\
\bottomrule
\end{tabular}
}
\caption{Results on AIME 24’, GPQA, and LCB (V1–4) using Llama-3.1-Nemotron-Nano-8B-v1. For both Length Scaling and Reasoning Memory we use the budget $m=30$ and $n=8$. }
\label{tab:reasoning-memory-vs-sft}
\end{table}

\subsection{Uncertainty Criteria}
\label{appendix-uncertainty-metrics}

Our main system uses generation length as the uncertainty proxy for stage 1 item-level filtering after retrieval. In this section, we compare this metric with several alternative criteria while keeping stage 2 fixed to length-based selection. Specifically, for each retrieved subroutine, we compute:

\begin{itemize}
\item \textbf{Likelihood}: log likelihood of the first 200 continued tokens conditioned on the retrieved subroutine. For this metric, lower is better as we expect the retrieval to inject new and unfamiliar information. 
\item \textbf{Entropy}: average token-level entropy of the first 200 continued tokens conditioned on the retrieved subroutine. Higher is better.
\item \textbf{Contrast}: the distributional difference between the downstream reasoning model and its base pretrained model, evaluated on the first 200 tokens generated in the continuation. This metric is inspired by \citet{DBLP:journals/corr/abs-2506-15710} and we instantiate it with KL Divergence (higher is better).
\item \textbf{Self-Eval}: a scalar score where the model is prompted to rate how relevant the retrieved subroutine is to the problem on a Likert scale. We repeatedly sample for 10 times with temperature = 0.6 and take the average as the score to apply the filter. 
\end{itemize}

All scores are then normalized using the same procedure as in \S\ref{section-approach-scaling}. Using DeepSeek-R1-Distill-Llama-8B with a budget of $m=30$, we report the task accuracy results in \Cref{tab:appendix-alt-metrics}. Overall, no single alternative dominates across all benchmarks, but length is the most reliable choice overall. On AIME 24’, entropy achieves the highest score (0.600), with length close behind at 0.575. On GPQA-D and LCB (V1–4), however, length clearly outperforms all other metrics (0.461 vs.\ 0.429 or lower on GPQA-D, and 0.310 vs.\ at most 0.273 on LCB). Averaged over the three tasks, length is therefore the strongest option, while also being significantly simpler and cheaper than metrics that require extra model evaluations. For this reason we adopt length as our default uncertainty proxy in the main experiments.

\begin{table}[t]
\centering
\begin{tabular}{lccc}
\toprule
\textbf{Metric} & \textbf{AIME 24’} & \textbf{GPQA-D} & \textbf{LCB (V1–4)} \\
\midrule
Likelihood & 0.571 & 0.429 & 0.273 \\
Entropy & \textbf{0.600} & 0.417 & 0.273 \\
Contrast & 0.563 & 0.419 & 0.269 \\
Self-Eval & 0.571 & 0.402 & 0.263 \\
Length & 0.575 & \textbf{0.461} & \textbf{0.310} \\
\bottomrule
\end{tabular}
\caption{\textbf{Alternative metrics for step 1 filtering.} Results for DeepSeek-R1-Distill-Llama-8B with budget m=30; stage 2 always uses length as the uncertainty proxy. Length achieves strong overall performance while being the simplest.}
\label{tab:appendix-alt-metrics}
\end{table}

\subsection{Qualitative Study}
\label{appendix-qualitative-study}

We conduct a small qualitative study in this section. \Cref{tab:qualitative-ex-aime2025-good,tab:qualitative-ex-math500-good,tab:qualitative-ex-livecode-good} show examples of Reasoning Memory runs with DeepSeek-R1-Distill-Llama-8B. Across examples, the self-generated query serves as a compact paraphrase of the model's latent goal (``convert between bases,'' ``factor a quadratic,'' etc.), which makes it well matched to the subquestions stored in the datastore. In all four examples, the retrieved subroutines are both relevant to the question and general enough to transfer across problems and benchmarks. We also observe that the continued reasoning rarely copies the hint verbatim. Instead, the model typically restates the problem and derives the key steps, sometimes referring to the retrieved subroutine later in the solution, suggesting that the hint acts more like a latent steering signal than an explicit template to fill in.

The two failure cases are shown in \Cref{tab:qualitative-ex-failure-1,tab:qualitative-ex-failure-2}. In the first failure, the self-verbalized query becomes overly specific, including particular numbers and surface details that occur infrequently in the datastore, which leads to irrelevant and thus less useful retrievals. In the second failure, the retrieved hint is physically reasonable but mismatched to the problem setup: the guidance correctly suggests using the Stefan--Boltzmann law but does not capture the problem-specific setup that the stars are blackbodies. As a result, the hint is generic and somewhat distracting. These cases highlight two current limitations of Reasoning Memory: query specificity and mismatch with the problem-specific setting.

\begin{table}[]
    \centering
    \begin{tabularx}{\linewidth}{@{}p{.11\linewidth}X@{}}
    \toprule
    \textit{Question} &
    Please answer the following math question. Find the sum of all integer bases $b > 9$ for which $17_b$ is a divisor of $97_b.$
    \\
    \midrule
    \textit{Query} &
    17 in base $b$ is equal to what in decimal?
    \\
    \midrule
    \textit{Retrieved Subquestion} &
    How do you convert a number from an arbitrary base $b$ to its decimal (base 10) equivalent?
    \\
    \midrule
    \textit{Associated Subroutine} &
    \begin{minipage}[t]{\linewidth}
    For problems like this, I should first write the number as a sum of each digit multiplied by the base raised to the power of its position index. Here's how:
    \begin{enumerate}
        \item \textbf{Identify each digit} of the number in the original base.
        \item \textbf{Assign positional indices} starting from $0$ at the rightmost digit (the least significant digit).
        \item \textbf{Compute the weight} of each digit by raising the base $b$ to the power of its position index.
        \item \textbf{Multiply each digit} by its corresponding weight.
        \item \textbf{Sum all the products} to obtain the decimal (base 10) equivalent.
    \end{enumerate}
    For example, a number $d_n d_{n-1} \ldots d_1 d_0$ in base $b$ converts to decimal as:
    \[
        d_n \cdot b^n + d_{n-1} \cdot b^{n-1} + \ldots + d_1 \cdot b^1 + d_0 \cdot b^0.
    \]
    Ensure all digits are valid in the original base (i.e., $0 \le d_i < b$) before proceeding.
    \end{minipage}
    \\
    \midrule
    \textit{Continued Reasoning} &
    \begin{minipage}[t]{\linewidth}
    Okay, so I have this math problem here: Find the sum of all integer bases $b > 9$ for which $17_b$ is a divisor of $97_b$. Hmm, okay. Let me try to figure this out step by step.

    First, I remember that when numbers are written with a subscript like $17_b$, it means they're in base $b$. So I need to convert both $17_b$ and $97_b$ into decimal (base 10) to work with them more easily.

    Let's start with $17_b$. In base $b$, the rightmost digit is the units place (which is $b^0$) and the next digit is the $b^1$ place. So $17_b$ would translate to:
    \[
        1 \times b^1 + 7 \times b^0 = b + 7.
    \]
    Got that. So $17_b = b + 7$ in decimal.

    Now, let's do the same for $97_b$…
    \end{minipage}
    \\
    \bottomrule
    \end{tabularx}
    \caption{An example from AIME 2025 where Reasoning Memory is effective.}
    \label{tab:qualitative-ex-aime2025-good}
\end{table}

\begin{table}[]
    \centering
    \begin{tabularx}{\linewidth}{@{}p{.11\linewidth}X@{}}
    \toprule
    \textit{Question} &
    Two-thirds of the students at Baker Middle School take music. There are 834 students who take music. How many students are there at Baker Middle School?
    \\
    \midrule
    \textit{Query} &
    $\tfrac{2}{3}$ of something equals 834, what is that something?
    \\
    \midrule
    \textit{Retrieved Subquestion} &
    What is the result of dividing 3.6 by 1.6?
    \\
    \midrule
    \textit{Associated Subroutine} &
    \begin{minipage}[t]{\linewidth}
    For problems like this, I should first eliminate decimals by expressing the division as a fraction of whole numbers. For example, if dividing \( a.b \) by \( c.d \), multiply both the numerator and denominator by 10 to convert them into whole numbers (\( \frac{ab}{cd} \)). Then, simplify the fraction by dividing the numerator and denominator by their greatest common divisor or perform long division directly. If simplifying isn’t straightforward, break the division into parts: calculate how many times the divisor fits into the dividend, subtract that portion, and handle the remainder to find the exact value. This method works for any decimal division where precision is needed.
    \end{minipage}
    \\
    \midrule
    \textit{Continued Reasoning} &
    \begin{minipage}[t]{\linewidth}
    Okay, so I've got this math problem here: Two-thirds of the students at Baker Middle School take music, and there are 834 students who take music. The question is asking how many students are there at Baker Middle School in total. Hmm, let's think about this step by step.

    Alright, two-thirds of the school takes music, which means if I can figure out what number multiplied by two-thirds gives me 834, that number will be the total number of students at the school. So, essentially, I need to solve for the total number of students, which I'll call \( S \).

    The equation would be:
    \[
        \frac{2}{3} \cdot S = 834.
    \]

    To find \( S \), I should isolate it on one side of the equation. To do that, I can multiply both sides by the reciprocal of two-thirds, which is three-halves \( \left(\frac{3}{2}\right) \). So, multiplying both sides by \( \frac{3}{2} \):

    \[
        S = 834 \cdot \frac{3}{2}.
    \]

    Let me compute that. \( 834 \) divided by \( 2 \) is \( 417 \), and then \( 417 \) multiplied by \( 3 \) is: \( 400 \times 3 = 1200 \), and \( 17 \times 3 = 51 \), so adding those together gives me \( 1251 \). So, \( S = 1251 \).
    \end{minipage}
    \\
    \bottomrule
    \end{tabularx}
    \caption{An example from MATH500 where Reasoning Memory is effective.}
    \label{tab:qualitative-ex-math500-good}
\end{table}

\begin{table}[]
    \centering
    \vspace{-4mm}
    \begin{tabularx}{\linewidth}{@{}p{.11\linewidth}X@{}}
    \toprule
    \textit{Question} &
    \begin{minipage}[t]{\linewidth}
    Generate a correct Python program that passes all tests for the given problem.\\[0.5em]
    \textbf{Problem Title:} $A x + B y < C$\\[0.5em]
    \textbf{Problem Statement:}\\
    You are given three length-$N$ sequences of positive integers:
    \[
      A = (A_1, A_2, \ldots, A_N),\quad
      B = (B_1, B_2, \ldots, B_N),\quad
      C = (C_1, C_2, \ldots, C_N).
    \]
    Find the number of pairs of positive integers $(x, y)$ that satisfy the following condition:
    \[
      A_i \times x + B_i \times y < C_i \quad \text{for all } 1 \leq i \leq N.
    \]
    It can be proved that the number of such pairs of positive integers satisfying the condition is finite.\\[0.25em]
    You are given $T$ test cases, each of which should be solved.
    \end{minipage}
    \\
    \midrule
    \textit{Query} &
    how to count the number of $(x,y)$ pairs that satisfy multiple inequalities $A_i x + B_i y < C_i$ for all $i$
    \\
    \midrule
    \textit{Retrieved Subquestion} &
    How do you count the number of integer pairs $(x, y)$ that satisfy a linear inequality with non-negative coefficients and bounds?
    \\
    \midrule
    \textit{Associated Subroutine} &
    \begin{minipage}[t]{\linewidth}
    For problems like this, I should consider the following steps:
    \begin{enumerate}
        \item \textbf{Fix one variable and solve for the other.} Choose one variable (e.g., $y$) and iterate through its possible values within the bounds. For each fixed $y$, determine the minimum or maximum value of the other variable (e.g., $x$) required to satisfy the inequality.
        \item \textbf{Use complementary counting if feasible.} If the inequality's valid region is large, it may be easier to calculate the total number of candidate pairs (for example, $(N+1)^2$ for bounds $0 \leq x, y \leq N$) and subtract the count of pairs that do \emph{not} satisfy the inequality. This is especially useful when the invalid region is smaller or has a simpler structure.
        \item \textbf{Account for variable bounds.} Ensure all computed ranges for $x$ or $y$ respect their given constraints (e.g., $0 \leq x \leq N$). Adjust the ranges to stay within these bounds, even if the inequality suggests values outside them.
        \item \textbf{Sum valid counts systematically.} For each fixed value of the first variable, compute the number of valid values for the second variable and accumulate these counts. Organize the calculations (for example, by grouping cases where the inequality trivially holds or fails) to avoid errors and redundant work.
    \end{enumerate}
    This approach balances direct enumeration with strategic shortcuts to efficiently count solutions while avoiding overcomplication.
    \end{minipage}
    \\
    \midrule
    \textit{Continued Reasoning} &
    \begin{minipage}[t]{\linewidth}
    So I've got this programming problem to solve. Let's read it carefully.

    The problem is about finding the number of pairs $(x, y)$ of positive integers that satisfy $A_i x + B_i y < C_i$ for all $i$ from $1$ to $N$. So for each $i$, the same $x$ and $y$ have to satisfy this inequality.

    In other words, we need to find all $(x, y)$ such that for \emph{every} $i$,
    \[
      A_i x + B_i y < C_i
    \]
    holds simultaneously. We are given $T$ test cases, and for each test case we get $N$ lines of $A_i, B_i, C_i$.

    The goal is: for each test case, compute the number of such $(x, y)$ pairs.

    Geometrically, each inequality $A_i x + B_i y < C_i$ describes a half-plane in the first quadrant (since $x$ and $y$ are positive integers). The feasible region is the intersection of all these half-planes restricted to $x \ge 1$, $y \ge 1$. The statement that the answer is finite means this intersection is a bounded polygonal region.

    One possible strategy is to treat $x$ as the main loop variable...
    
    \end{minipage}
    \\
    \bottomrule
    \end{tabularx}
    \caption{An example from LiveCodeBench where Reasoning Memory is effective.}
    \label{tab:qualitative-ex-livecode-good}
\end{table}

\begin{table}[]
    \centering
    \begin{tabularx}{\linewidth}{@{}p{.11\linewidth}X@{}}
    \toprule
    \textit{Question} &
    \begin{minipage}[t]{\linewidth}
    There are $n$ values of $x$ in the interval $0 < x < 2\pi$ where
    \[
        f(x) = \sin\bigl(7\pi \cdot \sin(5x)\bigr) = 0.
    \]
    For $t$ of these $n$ values of $x$, the graph of $y = f(x)$ is tangent to the $x$-axis. Find $n + t$.
    \end{minipage}
    \\
    \midrule
    \textit{Query} &
    How to find the number of solutions to $\sin(7\pi \cdot \sin(5x)) = 0$ between $0$ and $2\pi$, and how many of those have tangent to $x$-axis.
    \\
    \midrule
    \textit{Top-3 Retrieved Subquestion} &
    \begin{minipage}[t]{\linewidth}
    \begin{enumerate}
        \item How do you analyze the behavior of a trigonometric function (e.g., using derivatives) to confirm the number of solutions in a given interval?
        \item How do you systematically count and confirm the total number of valid solutions by combining results from both cases $(\sin(x) = 0$ and $2\cos(x) + \cos(2x) = 0)$?
        \item How do you analyze the behavior of a function (e.g., $f(x) = \tan(2x) + \sin(x)$) in intervals between vertical asymptotes to determine the number of solutions?
    \end{enumerate}
    \end{minipage}
    \\
    \bottomrule
    \end{tabularx}
    \caption{A failure example from AIME 2025 caused by over-specific queries.}
    \label{tab:qualitative-ex-failure-1}
\end{table}

\begin{table}[]
    \centering
    \begin{tabularx}{\linewidth}{@{}p{.11\linewidth}X@{}}
    \toprule
    \textit{Question} &
    \begin{minipage}[t]{\linewidth}
    % Astronomers are interested in the luminosities of two stars. Star 1 has a radius $1.5$ times larger than that of Star 2. The mass of Star 1 is also larger by the same factor ($1.5$ times). Through spectroscopy, the astronomers managed to determine the wavelengths at which the stars appeared brightest, and found that they are the same. Astronomers also calculated the radial velocities of the two stars to be $0$ and $700\ \mathrm{km/s}$ for Star 1 and Star 2 respectively. Assuming that the stars radiate as black bodies, they computed the luminosity of Star 1 to be greater than the luminosity of Star 2 by a factor of:\\[0.25em]
    % Choices:
    % \begin{itemize}
    %     \item[(A)] $\sim 2.32$
    %     \item[(B)] $\sim 2.23$
    %     \item[(C)] $\sim 2.35$
    %     \item[(D)] $\sim 2.25$
    % \end{itemize}
    [redacted, luminosities of two stars radiating as black bodies]
    \end{minipage}
    \\
    \midrule
    \textit{Query} &
    astronomers luminosity star radius relation
    \\
    \midrule
    \textit{Retrieved Subquestion} &
    What is the relationship between a star's radius and its luminosity?
    \\
    \midrule
    \textit{Associated Subroutine} &
    \begin{minipage}[t]{\linewidth}
    For problems asking about the relationship between a star's radius and luminosity, I should start by recalling the Stefan--Boltzmann law:
    \[
        L = 4\pi R^2 \sigma T^4,
    \]
    where $L$ is luminosity, $R$ is radius, $T$ is temperature, and $\sigma$ is the Stefan--Boltzmann constant. Luminosity depends on both radius and temperature. To isolate the radius--luminosity relationship, consider how changes in radius affect $L$ while holding temperature constant. However, for main-sequence stars, radius and temperature are not independent---both increase with mass. Use the mass--radius and mass--luminosity relationships (e.g., $L \propto M^{3.5}$, $R \propto M^{0.7}$) to deduce the overall correlation. Remember that temperature also contributes strongly to luminosity, so the radius’s role must be analyzed in concert with mass and temperature dependencies.
    \end{minipage}
    \\
    \midrule
    \textit{Continued Reasoning} &
    \begin{minipage}[t]{\linewidth}
    Okay, so I have this astronomy problem here about the luminosities of two stars, Star 1 and Star 2. Let me try to work through it step by step. The question says that Star 1 has a radius $1.5$ times larger than Star 2, and the mass of Star 1 is also $1.5$ times larger. They also mention that the wavelengths at which the stars appear brightest are the same, and their radial velocities are $0$ for Star 1 and $700\ \mathrm{km/s}$ for Star 2. The question is about the ratio of their luminosities, so I need to compute how much brighter Star 1 is compared to Star 2.

    I remember that the luminosity of a star is related to its radius and temperature. The Stefan--Boltzmann law is
    \[
        L = 4\pi R^2 \sigma T^4,
    \]
    where $L$ is luminosity, $R$ is radius, $\sigma$ is the Stefan--Boltzmann constant, and $T$ is the effective temperature. At first I might wonder whether the Doppler effect due to different radial velocities matters here, but the problem states that the wavelengths at which they appear brightest are the same. That suggests their effective temperatures are the same, because the peak wavelength (or color) is determined by temperature via Wien's law. So I can take $T_1 = T_2$.

    If the temperatures are equal, then the ratio of luminosities depends only on the ratio of radii:
    \[
        \frac{L_1}{L_2} = \left(\frac{R_1}{R_2}\right)^2 \left(\frac{T_1}{T_2}\right)^4 = \left(\frac{R_1}{R_2}\right)^2.
    \]
    Since $R_1 = 1.5\,R_2$, this gives
    \[
        \frac{L_1}{L_2} = (1.5)^2 = 2.25,
    \]
    which corresponds to choice (D), approximately $2.25$.
    \end{minipage}
    \\
    \bottomrule
    \end{tabularx}
    \caption{A failure example from GPQA-D caused by problem setup mismatch.}
    \label{tab:qualitative-ex-failure-2}
\end{table}

\end{document}